\documentclass[11pt,letterpaper]{article}
\usepackage{emnlp2016}
\usepackage{times}
\usepackage{latexsym}
\usepackage{amsfonts}
\usepackage{graphicx}
\usepackage{verbatim}
\usepackage{bm}
\usepackage{amsmath}
\usepackage{subfig}
\usepackage{threeparttable}% http://ctan.org/pkg/threeparttable
\usepackage{booktabs}% http://ctan.org/pkg/booktabs
\usepackage{tabularx}% http://ctan.org/pkg/tabularx
\usepackage{multirow, array, hhline}
\usepackage{caption}
\usepackage{url}
\usepackage[labelfont=normal]{caption}%http://ctan.org/pkg/caption
\emnlpfinalcopy

\title{Rationale-Augmented Convolutional Neural Networks\\ for Text Classification}

\author{Ye Zhang,$^1$~
  Iain Marshall$,^2$~
  Byron C. Wallace$^3$\\
  $^1$Department of Computer Science, University of Texas at Austin\\
  $^2${Department of Primary Care and Public Health Sciences, King’s College London}\\
  $^3${College of Computer and Information Science, Northeastern University}\\
  {\tt yezhang@cs.utexas.edu,  iain.marshall@kcl.ac.uk} \\
       {\tt  byron@ccs.neu.edu }
 }

\date{}

\begin{document}

\maketitle

% bcw: as usual, probably too long!
\begin{abstract}

We present a new Convolutional Neural Network (CNN) model for text classification that jointly exploits labels on documents and their constituent sentences. Specifically, we consider scenarios in which annotators explicitly mark sentences (or snippets) that support their overall document categorization, i.e., they provide \emph{rationales}. Our model exploits such supervision via a hierarchical approach in which each document is represented by a linear combination of the vector representations of its component sentences. We propose a sentence-level convolutional model that estimates the probability that a given sentence is a rationale, and we then scale the contribution of each sentence to the aggregate document representation in proportion to these estimates. Experiments on five classification datasets that have document labels and associated rationales demonstrate that our approach consistently outperforms strong baselines. Moreover, our model naturally provides explanations for its predictions.  % Exploiting rationales can improve the performance of CNNs for text classification by up to $\sim$5\% in absolute accuracy while . %This is the first work to incorporate rationales into neural models for text classification. 

\end{abstract}

\section{Introduction}
\vspace{-.25em} 

% bcw: maybe cite the yoav survey for below? or socher? 
Neural models that exploit word embeddings have recently achieved impressive results on text classification tasks \cite{goldberg2015primer}. Feed-forward Convolutional Neural Networks (CNNs), in particular, have emerged as a relatively simple yet powerful class of models for text classification \cite{kim2014convolutional}.%,zhang2015}.  %johnson2014effective,zhang2015character,zhang2016mgnc, % bcw: we cite these later anyway

%In this work we focus on the task of document classification. 
These neural text classification models have tended to assume a standard supervised learning setting in which instance labels are provided. Here we consider an alternative scenario in which we assume that we are provided a set of \emph{rationales}
\cite{zaidan2007using,zaidan2008,McDonnell16-hcomp} in addition to instance labels, i.e., sentences or snippets that support the corresponding document categorizations. Providing such rationales during manual classification is a natural interaction for annotators, and requires little additional effort \cite{settles2011,McDonnell16-hcomp}. Therefore, when training new classification systems, it is natural to acquire supervision at both the document and sentence level, with the aim of inducing a better predictive model, potentially with less effort. 

%were built on top of sparse representations for text classification, and linear models such as Support Vector Machines (SVMs)~\cite{joachims1998text}. % bcw: maybe cite vapnik??

%The intuition we will exploit is that it will often be the case that only a handful of sentences in any given document are likely to inform its classification with respect to a given categorization target. F

%ollowing previous work \cite{zaidan2007using,zaidan2008}, we refer to these sentences (or snippets) that are relevant to the classification task at hand as \emph{rationales}. 

%can play a role in determining the label of the document, while other sentences are less important. We call such sentence `rationales'. It's necessary to explore how to utilize these rationales during training to help with the classification tasks. 

% bcw notes to self (5/2)
% 1 mention that this helps with interpretability because we **predict** rationales naturally
% 2 emphasize that our main contribution here is an augmentation to CNN that exploits rationales

Learning algorithms must be designed to capitalize on these two types of supervision. Past work (Section \ref{section:related-work}) has introduced such methods, but these have relied on linear models such as Support Vector Machines (SVMs)~\cite{joachims1998text}, operating over sparse representations of text. We propose a novel CNN model for text classification that exploits both document labels and associated rationales. %We demonstrate that our model substantially outperforms relevant baselines, including previously proposed models that capitalize on rationales \cite{zaidan2007using,marshall2016robotreviewer} and multiple baseline CNN variants, including a CNN equipped with an attention mechanism. %Our model uniformly outperforms these baseline approaches, and we achieve state-of-the-art performance with respect to classifying articles describing the conduct and results of clinical trials as being at high or low risk of bias. 

Specific contributions of this work as follows. (1) This is the first work to incorporate rationales into neural models for text classification. (2) Empirically, we show that the proposed model uniformly outperforms relevant baseline approaches across five datasets, including previously proposed models that capitalize on rationales \cite{zaidan2007using,marshall2016robotreviewer} and multiple baseline CNN variants, including a CNN equipped with an attention mechanism. We also report state-of-the-art results on the important task of automatically assessing the risks of bias in the studies described in full-text biomedical articles \cite{marshall2016robotreviewer}. (3) Our model naturally provides \emph{explanations} for its predictions, providing interpretability.

%Ye_13: implementation !!!!
% bcw_14: excellent. just making new paragraph.
We have made available online both a Theano\footnote{\url{https://github.com/yezhang-xiaofan/Rationale-CNN}} and a Keras implementation\footnote{\url{https://github.com/bwallace/rationale-CNN}} of our model.

%(3) Our model achieves state-of-the-art results on the task of \emph{risk-of-bias} assessment \cite{marshall2016robotreviewer}, which involves classifying the results reported in articles describing the conduct and results of clinical trials as being at low or high risk of statistical bias. 

% bcw: can we claim state of the art on movies dataset??
%Ye: Can we claim state of the art on RoB dataset?? We don't try other neural models. 

%We develop a document classification system called Dual Supervision Convolution Neural Network (RA-CNN), where we have information on which sentences in the document are rationales and support the label of the document.  In other words, in the training set, we not only have labels of documents, but we also have labels of each sentence. In RA-CNN, we first use a one-layer CNN to represent each sentence in the document, and train this CNN using the sentence labels. Intuitively, this model will learn parameters that highlight the rationals. Then we sum the sentence vector up weighted by the probability of each sentence being the rationale, to obtain the document vector representation. We then use the document label to train a classifier. The key point is that the sentence and document representation share the same word embedding and convolution layers, so our system can jointly learn parameters that can support rationales, which in turn helps the classification on documents. 

\section{Related Work}
\label{section:related-work}

%\noindent {\bf Neural models for text classification}.

\subsection{Neural models for text classification}
\newcite{kim2014convolutional} proposed the basic CNN model we describe below and then build upon in this work. Properties of this model were explored empirically in \cite{zhang2015}. We also note that \newcite{zhang2016mgnc} extended this model to jointly accommodate multiple sets of pre-trained word embeddings. Roughly concurrently to Kim, \newcite{johnson2014effective} proposed a similar CNN architecture, although they swapped in one-hot vectors in place of (pre-trained) word embeddings. They later developed a semi-supervised variant of this approach \cite{johnson2015semi}.  

In related recent work on Recurrent Neural Network (RNN) models for text, \newcite{tang2015document} proposed using a Long Short Term Memory (LSTM) layer to represent each sentence and then passing another RNN variant over these. And \newcite{yang2016hierarchical} proposed a hierarchical network with two levels of attention mechanisms for document classification. We discuss this model specifically as well as attention more generally and its relationship to our proposed approach in Section \ref{section:AT-CNN}. 

%For example,

%\vspace{.35em}
%\noindent {\bf Exploiting rationales}. 

\subsection{Exploiting \emph{rationales}}
%13: Reviewer 1 asked to cite ko2002automatic. I also add some other citations. 
In long documents the importance of sentences varies; some are more central than others. Prior work has investigated methods to measure the relative importance sentences~\cite{ko2002automatic,murata2000japanese}. In this work we adopt a particular view of sentence importance in the context of document classification. In particular, we assume that documents comprise sentences that directly support their categorization. We call such sentences \emph{rationales}.

%In this work we assume that within a given document there exist sentences labeled as supportive for the label of the document, which is called rationale. 
The notion of rationales was first introduced by \newcite{zaidan2007using}. To harness these for classification, they proposed modifying the Support Vector Machine (SVM) objective function to encode a preference for parameter values that result in instances containing manually annotated rationales being more confidently classified than `pseudo'-instances from which these rationales had been stripped. This approach dramatically outperformed baseline SVM variants that do not exploit such rationales. \newcite{yessenalina2010automatically} later developed an approach to \emph{generate} rationales. %~\cite{yessenalina2010multi} jointly learn the rationales and document labels. 
% bcw_14: i didn't get it -- what does it mean to learn rationales? as in, unsupervised? 
%Ye_14: Yes. There is no supervision on sentences. The model is trained only on document labels directly. And it simultaneously extracts useful sentences.   

%and proposes a new Support Vector Machine (SVM) framework that can incorporate rationales information during training. Their model greatly boosts the performance compared with situations without rationales. 
Another line of related work concerns models that capitalize on \emph{dual supervision}, i.e., labels on individual \emph{features}. This work has largely involved inserting constraints into the learning process that favor parameter values that align with \emph{a priori} feature-label affinities or rankings \cite{druck2008learning,mann2010generalized,small2011constrained,settles2011}. We do not discuss this line of work further here, as our focus is on exploiting provided rationales, rather than individual labeled features.

%For example, \newcite{settles2011} proposed a fast, simple approach based on Naive Bayes that adjusts priors to reflect feature labels. Similarly, \newcite{small2011constrained} proposed the Constrained Weighed SVM (CS-SVM) model, which constrained parameter estimation to prefer weight vectors that agreed with \emph{a priori} feature rankings provided by domain experts. Finally, \newcite{druck2008learning,mann2010generalized} proposed a general approach to learning from feature supervision via soft constraints which they termed \emph{Generalized Expectation Criteria}. We do not discuss this line of work further here, as our focus in this work is on exploiting provided rationales, rather than labeled features.

%The first work to popularize Convolutional Neural Networks (CNNs) for text (specifically, sentence) classification was \newcite{kim2014convolutional}. We review the basic architecture in Section \ref{section:preliminaries}. 

%For sentence classification, \cite{kim2014convolutional} uses a one-layer Convolution Neural Network (CNN) model with pre-trained word embedding. 

% bcw (notes to self)
% should cite settles DUALIST paper (EMNLP i think?)
% should also cite prem melville's stuff
% need to be clear that rationales different than just labeled features
% 

\section{Preliminaries: CNNs for text classification}
\label{section:preliminaries} 

% bcw 5/3 should we add the final (output) nodes above o? 
\begin{figure}[ht!]
\centering
\includegraphics[width=0.35\textwidth]{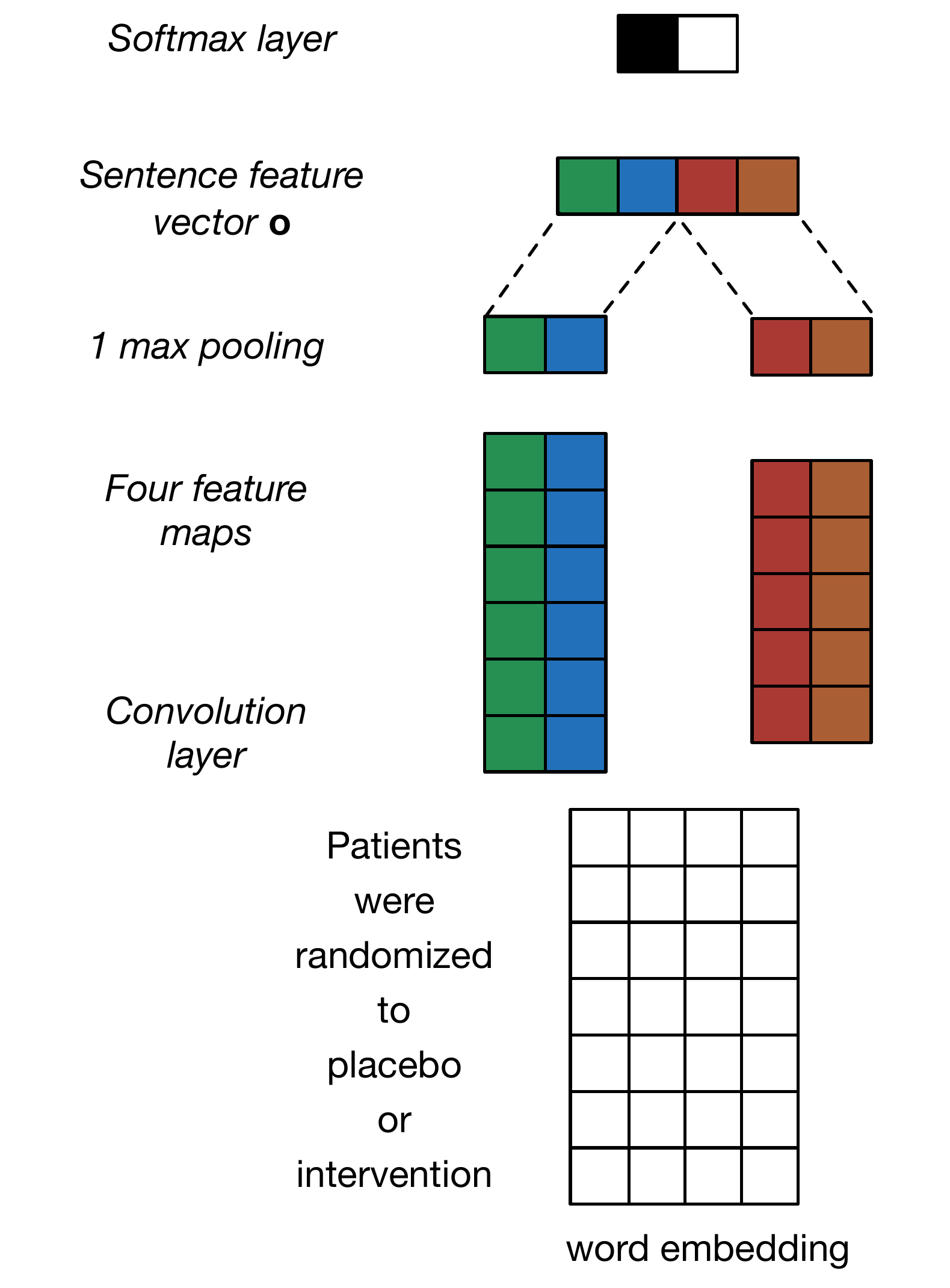}
\caption{A toy example of a CNN for sentence classification. Here there are four filters, two with heights 2 and two with heights 3, resulting in feature maps with lengths 6 and 5 respectively.}
\label{Single_CNN}
\end{figure}

%In this Section we review preliminaries to contextualize the novel model we propose Section \ref{section:ra-CNN}. 

%\subsection{Basic CNN}
%\label{section:standard-CNN}

% bcw: probably should allude to where such pre-trained embeddings would come from... i know everyone probably knows, but still...
%Ye: I mentioned it in the experimental setting, still mention here? 
%Ye_09: I add more details in case reviewer 1 does not understand 
We first review the simple one-layer CNN for sentence modeling proposed by \newcite{kim2014convolutional}. Given a sentence or document comprising $n$ words $w_1,w_2,$...,$w_n$, we replace each word with its $d$-dimensional pretrained embedding, and stack them row-wise, generating an instance matrix $A\in\mathbb{R}^{n\times d}$.

We then apply convolution operations on this matrix using multiple linear filters, these will have the same width $d$ but may vary in \emph{height}. Each filter thus effectively considers distinct $n$-gram features, where $n$ corresponds to the filter height. In practice, we introduce multiple, redundant features of each height; thus each filter height might have hundreds of corresponding instantiated filters. Applying filter $i$ parameterized by $W_i\in\mathbb{R}^{h_i\cdot d}$ to the instance matrix induces a \emph{feature map} $f_i\in\mathbb{R}^{n-h_i+1}$. This process is performed by sliding the filter from the top of the matrix (the start of the document or sentence) to the bottom. At each location, we apply element-wise multiplication between filter $i$ and sub-matrix $A[j:j+h_i-1]$, and then sum up the resultant matrix elements. In this way, we induce a vector (feature map) for each filter. 

We next run the feature map through an element-wise non-linear transformation. Specifically, we use the \emph{Rectified Linear Unit}, or ReLU~\cite{krizhevsky2012imagenet}. We extract the maximum value $o_i$ from each feature map $i$ (\emph{1-max pooling}).

Finally, we concatenate all of the features $o_i$ to form a vector representation $\mathbf{o}\in\mathbb{R}^{|F|}$ for this instance, where $|F|$ denotes the total number of filters. Classification is then performed on top of $\mathbf{o}$, via a softmax function. Dropout~\cite{srivastava2014dropout} is often applied at this layer as a means of regularization. We provide an illustrative schematic of the basic CNN architecture just described in Figure \ref{Single_CNN}. For more details, see~\cite{zhang2015}.

% bcw 5/2: not crazy about the name sentCNN; this actually makes it sound (to me) like it takes into account sentence structure, when the opposite is true! I suggest perhaps just referring to this as CNN, since it's the most standard variant?
This model was originally proposed for sentence classification \cite{kim2014convolutional}, but we can adapt it for document classification by simply treating the document as one long sentence. We will refer to this basic CNN variant as \textbf{\textit{CNN}} in the rest of the paper. Below we consider extensions that account for document structure.
%In this way, we obtain a vector representation for the entire document, then we use a softmax function on this vector to do classification. 

\section{Rationale-Augmented CNN for Document Classification}
\label{section:ra-CNN}

We now move to the main contribution of this work: a rationale-augmented CNN for text classification. We first introduce a simple variant of the above CNN that models document structure (Section \ref{section:doc-CNN}) and then introduce a means of incorporating rationale-level supervision into this model (Section \ref{section:ra-cnn-ss}). In Section \ref{section:AT-CNN} we discuss connections to attention mechanisms and describe a baseline equipped with one, inspired by \newcite{yang2016hierarchical}.  

%% bcw: I like this point, but this should be moved 
% TODO add back in in eval section!
% just choose a simple but strong CNN model and we won't compare CNN and Doc-CNN with other already existed or more complicated vector representation methods. 
%\footnote{We actually found that simple CNN model perform much better than (hierachical) LSTM and GRU model on our datasets, and CNN is relatively more robust and less sensitive to hyper-parameters compared to LSTM and GRU.}

%% bcw 5/2: this *was* above, but I think this is part of our model. It's worth noting that this specific architecture (although simple!) hasn't been proposed before, right? or has it? 
%Ye: Doc-CNN has been used widely, though always as a baseline. 
\subsection{Modeling Document Structure}
\label{section:doc-CNN}

% bcw: needs math! i think best to write out the model explicitly
Recall that rationales are snippets of text marked as having supported document-level categorizations. We aim to develop a model that can exploit these annotations during training to improve classification. Here we achieve this by developing a hierarchical model that estimates the probabilities of individual sentences being rationales and uses these estimates to inform the document level classification.

%Here we achieve this by explicitly accounting for the probabilities of a document's constituent sentences being rationales

%informing the document level classification of the probabilities of its constituent sentences being relevant to this.

As a first step, we extend the CNN model above to explicitly account for document structure. Specifically, we apply a CNN to each individual sentence in a document to obtain sentence vectors independently. We then sum the respective sentence vectors to create a document vector.\footnote{We also experimented with taking the average of sentence vectors, but summing performed better in informal testing.
} 
As before, we add a softmax layer on top of the document-level vector to perform classification. We perform regularization by applying dropout both on the individual sentence vectors and the final document vector. We will refer to this model as \textbf{\textit{Doc-CNN}}. Doc-CNN forms the basis for our novel approach, described below.

%Note that this is in contrast to the CNN architecture reviewed above, in which sentence structure was ignored.

%% bcw: notes on the estimation should be moved elsewhere 
%For both CNN and Doc-CNN, we'll use back-propagation algorithm~\cite{lecun2015deep} and train the word embedding layer, the convolution layer and the softmax layer. 

\subsection{RA-CNN} 
\label{section:ra-cnn-ss} 

% bcw: as per my email, I think we should write the model out. for example, actually write out the weighted sentence level aggregation function

\begin{figure*}[t]
\includegraphics[width=1.0\textwidth]{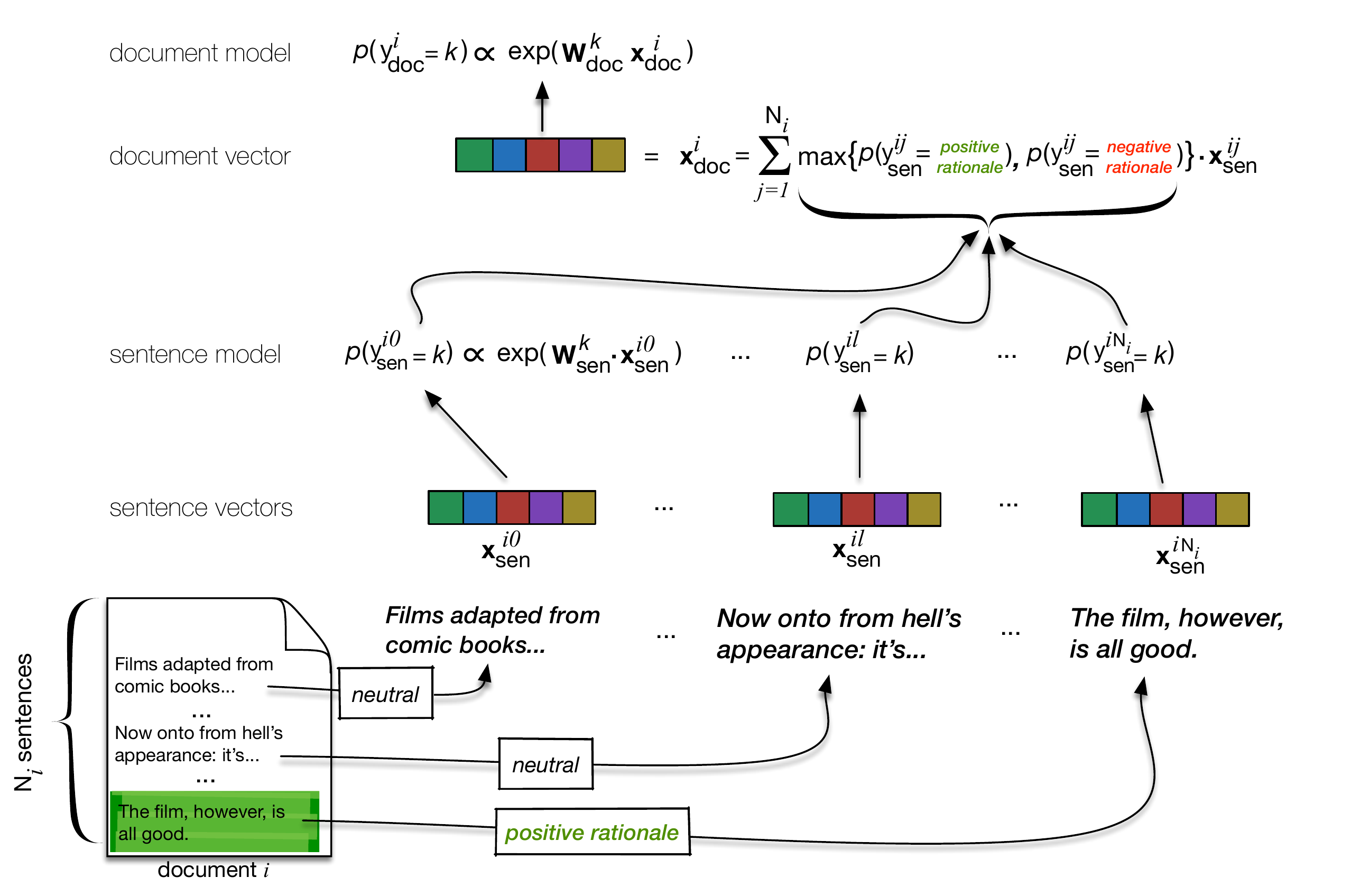}
\caption{A schematic of our proposed Rationale-Augmented Convolution Neural Network (RA-CNN). The sentences comprising a text are passed through a sentence model that outputs probabilities encoding the likelihood that sentences are neutral or a (positive or negative) rationale. Sentences likely to be rationales are given higher weights in the global document vector, which is the input to the document model.}
\label{DS_CNN}
\end{figure*}

In this section we present the Rationale-Augmented CNN (\textit{\textbf{RA-CNN}}). Briefly, RA-CNN induces a document-level vector representation by taking a \emph{weighted} sum of its constituent sentence vectors. Each sentence weight is set to reflect the estimated probability that it is a rationale in support of the most likely class. We provide a schematic of this model in Figure \ref{DS_CNN}. 

% bcw 5/3: i thought it kind of confusing to denote the vectors by CNN so changed to x.
RA-CNN capitalizes on both sentence- and document-level supervision. There are thus two steps in the training phase: \emph{sentence level} training and \emph{document level} training. For the former, we apply a CNN to each sentence $j$ in document $i$ to obtain sentence vectors $\mathbf{x}_\text{sen}^{ij}$. We then add a softmax layer parametrized by $\textbf{W}_\text{sen}$; this takes as input sentence vectors. We fit this model to maximize the probabilities of the observed rationales:

%  using the sentence labels ($y_{\text{sen}}^{ij}$), 
%estimating $\textbf{E},\textbf{C}$ and $\textbf{W}_\text{sen}$ to maximize the probabilities of the observed rationales:

\begin{equation}
%\vspace{-1em}
   p(y_{\text{sen}}^{ij}=k; \textbf{E},\textbf{C},\textbf{W}_\text{sen}) = \frac{\text{exp}(\textbf{W}_\text{sen}^{(k)T}\mathbf{x}_{\text{sen}}^{ij})}{\sum_{k=1}^{K_\text{sen}}\text{exp}(\textbf{W}_\text{sen}^{(k)T}\mathbf{x}_{\text{sen}}^{ij})}
   \label{Eq:sen}
\end{equation}

\noindent Where $y_{\text{sen}}^{ij}$ denotes the rationale label for sentence $j$ in document $i$, $K_\text{sen}$ denotes the number of possible classes for sentences, \textbf{E} denotes the word embedding matrix, \textbf{C} denotes the convolution layer parameters, and $\textbf{W}_\text{sen}$ is a matrix of weights (comprising one weight vector per sentence class).%The corresponding loss function we aim to minimize is the categorical cross-entropy on sentence labels. 

% bcw: putting back in for now; can cut if we need space later!
\begin{comment}
\begin{equation}
    J(\textbf{E},\textbf{C},\textbf{W}_\text{sen}) = -\sum_{i=1}^{B}\sum_{k=1}^K 1\{y_\text{sen}^{ij}=k\}\text{log}p(y_\text{sen}^{ij}=k)
    \label{Eq:sen-loss}
\end{equation}
\end{comment}
%Ye: This equation is confusing !
% bcw: fine, fine

In our setting, each sentence has three possible labels ($K_\text{sen}=3$). When a rationale sentence appears in a positive document,\footnote{All of the document classification tasks we consider here are binary, although extension of our model to multi-class scenarios is straight-forward.} it is a \textit{positive rationale}; when a rationale sentence appears in a negative document, it is a \textit{negative rationale}. All other sentences belong to a third, neutral class: these are \textit{non-rationales}. 
%Ye_13: I move the following footnote to the text, since two reviewers did not see this in their reviews. 
We also experimented with having only two sentence classes: \emph{rationales} and \emph{non-rationales}, but this did not perform as well as explicitly maintaining separate classes for rationales of different polarities.

% bcw: write objectives down
%After we finish training on sentences in the training set, we can get an estimator for sentences using Eq. (\ref{Eq:sen}).

We train an estimator using the provided rationale annotations, optimizing over \{$\textbf{E},\textbf{C},\textbf{W}_\text{sen}$\} to minimize the categorical cross-entropy of sentence labels. Once trained, this sub-model can provide conditional probability estimates regarding whether a given sentence is a positive or a negative rationale, which we will denote by $p_\text{pos}$ and $p_\text{neg}$, respectively. %We consider the strength $\text{max}\{p_\text{pos}$, $p_\text{neg}\}
%It then follows that the probability of a sentence being a \textit{non-rationale} is 1-$p_\text{pos}$-$p_\text{neg}$.

%, where weights are set to the predicted probabilities that the corresponding sentences are rationales. More precisely, 
We next train the document-level classification model. The inputs to this are vector representations of documents, induced by summing over constituent sentence vectors, as in Doc-CNN. However, in the RA-CNN model this is a weighted sum. Specifically, weights are set to the estimated probabilities that corresponding sentences are rationales in the most likely direction. More precisely:
%\vspace{-0.4in}
\begin{equation}
    \mathbf{x}_{\text{doc}}^i = \sum_{j=1}^{N_i}\mathbf{x}_{\text{sen}}^{ij} \cdot \text{max}\{p_{\text{pos}}^{ij},p_{\text{neg}}^{ij}\}
\end{equation}

\noindent Where $N_i$ is the number of sentences in the $i$th document.
The intuition is that sentences likely to be rationales will have greater influence on the resultant document vector representation, while the contribution of neutral sentences (which are less relevant to the classification task) will be minimized. 

The final classification is performed by a softmax layer parameterized by $\mathbf{W}_\text{doc}$; the inputs to this layer are the document vectors. The $\mathbf{W}_\text{doc}$ parameters are trained using the document-level labels, $y_\text{doc}^i$:

\begin{equation}
    p(y_\text{doc}^i=k;\mathbf{E},\mathbf{C},\mathbf{W}_\text{doc}) = \frac{\text{exp}(\mathbf{W}_\text{doc}^{(k)T}\mathbf{x}_{\text{doc}}^{i})}{\sum_{k=1}^{K_\text{doc}}\text{exp}(\mathbf{W}_\text{doc}^{(k)T}\mathbf{x}_{\text{doc}}^i)}
\end{equation}

\noindent where $K_{\text{doc}}$ is the cardinality of the document label set. We optimize over parameters to minimize cross-entropy loss (w.r.t. the document labels). %We use the cross entropy as the loss function:

% bcw: 5/10 -- cutting for space for now.

% bcw: 5/3 -- ok probably don't need to produce this loss twice after all
\begin{comment}
\begin{align*}
    J(\mathbf{E},\mathbf{C},\mathbf{W}_\text{doc}) = -\sum_{i=1}^{M}\sum_{k=1}^K 1\{y_\text{doc}^i=k\}\\
    *\text{log}p(y_\text{doc}^i=k)
\end{align*}
\end{comment} 

%where $M$ is the number of documents in one mini-batch. 
% Ye: but this loss is different from the sentence level loss. 

We note that the sentence- and document-level models share word embeddings $\mathbf{E}$ and convolution layer parameters $\mathbf{C}$, but the document-level model has its own softmax parameters $\mathbf{W}_\text{doc}$. When training the document-level model, $\mathbf{E}$, $\mathbf{C}$ and $\mathbf{W}_\text{doc}$ are fit, but we hold $\mathbf{W}_\text{sen}$ fixed. 

%To generate a document vector at test time, we take a weighted sum of its constituent sentence vectors, where sentence weights encode the predicted probabilities of corresponding sentences being rationales. We then run this document-vector through the final softmax to generate a prediction.

%and use the document level classifier to classify the document vector into positive or negative. We apply dropout on the document vector as well. 

The above two-step strategy can be equivalently described as follows. We first estimate $\mathbf{E}$, $\mathbf{C}$ and $\mathbf{W}_\text{sen}$, which parameterize our model for identifying rationales in documents. We then move to fitting our document classification model. For this we initialize the word embedding and convolution parameters to the $\mathbf{E}$ and $\mathbf{C}$ estimates from the preceding step. We then directly minimize the document level classification objective, tuning $\mathbf{E}$ and $\mathbf{C}$ and simultaneously fitting $\mathbf{W}_\text{doc}$. 

Note that this sequential training strategy differs from the \emph{alternating} training approach commonly used in multi-task learning~\cite{collobert2008unified}. We found that the latter approach does not work well here, leading us to instead adopt the cascade-like feature learning approach~\cite{collobert2008unified} just described. %One may view our strategy as \emph{pre-training} sentence representations, which are subsequently used during document classification. 

\begin{comment}
In our case, it's quite different from multi-task training, since we only have one final task: classifying documents, and classifying sentences into whether a rationale or not is just one intermediate step. We treat it as a pre-training task before document classification, and experimentally, it performs better than alternative training.
\end{comment}

One nice property of our model is that it naturally provides explanations for its predictions: the model identifies rationales and then categorizes documents informed by these. Thus if the model classifies a test instance as \emph{positive}, then by construction the sentences associated with the highest $p_{\text{pos}}^{ij}$ estimates are those that the model relied on most in coming to this disposition. These sentences can of course be output in conjunction with the prediction. We provide concrete examples of this in Section \ref{section:predicted-rationales}.

\subsection{Rationales as `Supervised Attention'}
\label{section:AT-CNN}

%An alternative to our RA-CNN is not to use sentence level supervision. 
One may view RA-CNN as a supervised variant of a model equipped with an \emph{attention mechanism} \cite{bahdanau2014neural}. On this view, it is apparent that rather than capitalizing on rationales directly, we could attempt to let the model learn which sentences are important, using \emph{only} the document labels. We therefore construct an additional baseline that does just this, thereby allowing us to assess the impact of learning directly from rationale-level supervision. 

%attention mechanism built on sentence vectors similar to the work in~\cite{yang2016hierarchical}, to let the model learn which sentence is more worth noting by just using document labels. 

Following the recent work of \newcite{yang2016hierarchical}, we first posit for each sentence vector a hidden representation $\mathbf{u}^{ij}_{\text{sen}}$. We then define a sentence-level \emph{context vector} $\mathbf{u}_s$, which  
%Ye_09: according to the comment "vectors can be multiplied but not applied"
we multiply with each $\mathbf{u}^{ij}_{\text{sen}}$ to induce a weight $\alpha_{ij}$. Finally, the document vector is taken as a weighted sum over sentence vectors, where weights reflect $\alpha$'s. We have:

\begin{equation}
    \mathbf{u}^{ij}_{\text{sen}} = \text{tanh}(\mathbf{W}_s\mathbf{x}^{ij}_{\text{sen}}+b_s)
\end{equation}
\begin{equation}
    \alpha_{ij} = \frac{\text{exp}(\mathbf{u}_s^T \mathbf{u}^{ij}_{\text{sen}})}{\sum_j^{N_i} \text{exp}(\mathbf{u}_s^T\mathbf{u}^{ij}_{\text{sen}})}
\end{equation}

\begin{equation}
    \mathbf{x}_{\text{doc}}^i = \sum_j^{N_i}\alpha_{ij}\mathbf{x}_{\text{sen}}^{ij}
\end{equation}
where $\mathbf{x}_{\text{doc}}^i$  again denotes the document vector fed into a softmax layer, and $\mathbf{W}_s$, $\mathbf{u}_s$ and $b_s$ are learned during training. We will refer to this attention-based method as \emph{AT-CNN}.%, and we include this as one of our baselines.%, and use it as one of the baselines. 

% bcw: I don't think we explicitly need to say it's one of our baselines here

\section{Datasets}
\label{section:datasets}

We used five text classification datasets to evaluate our approach in total. Four of these are biomedical text classification datasets (\ref{section:risk-of-bias-data}) and the last is a collection of movie reviews (\ref{section:movie-review-dataset}). These datasets share the property of having recorded rationales associated with each document categorization. We summarize attributes of all datasets used in this work in Table \ref{Dataset}.

\subsection{Risk of Bias (RoB) Datasets}
\label{section:risk-of-bias-data}

% bcw: iain please make sure you agree with my brief description!!!
We used a collection \emph{Risk of Bias} (RoB) text classification datasets, described at length elsewhere~\cite{marshall2016robotreviewer}. Briefly, the task concerns assessing the reliability of the evidence presented in full-text biomedical journal articles that describe the conduct and results of randomized controlled trials (RCTs). This involves, e.g., assessing whether or not patients were properly blinded as to whether they were receiving an active treatment or a comparator (such as a placebo). If such blinding is not done correctly, it compromises the study by introducing statistical bias into the treatment efficacy estimate(s) derived from the trial. 

A formal system for making bias assessments is codified by the Cochrane Risk of Bias Tool \cite{higgins2011cochrane}. This tool defines multiple \emph{domains}; the risk of bias may be assessed in each of these. We consider four domains here. (1) Random sequence generation (RSG): were patients were assigned to treatments in a truly random fashion? (2) Allocation concealment (AC): were group assignments revealed to the person assigning patients to groups (so that she may have knowingly or unknowingly) influenced these assignments? (3) Blinding of Participants and Personnel (BPP): were all trial participants and individuals involved in running the trial blinded as to who was receiving which treatment? (4) Blinding of outcome assessment (BOA): were the parties who measured the outcome(s) of interest blinded to the intervention group assignments? These assessments are somewhat subjective. To increase transparency, researchers performing RoB assessment therefore record rationales (sentences from articles) supporting their assessments.  %These sentences constitute our rationales for the RoB datasets.

\begin{comment}
The task is to identify whether a clinical trial is bias. 
RoB dataset has four domains: Random sequence generation (RSG), Allocation concealment (AC), Blinding of participants and personnel (BPP) and Blinding of outcome assessment (BOA). We treat these four domains as four independent datasets, so equivalently, we have four datasets. We'll run 5-fold cross validation (CV) on these four datasets. 
\end{comment}

\begin{table}%[ht]
\centering
\small
 \begin{tabular}{c c c c c} 
 %\hline
& \emph{N} &\emph{\#sen}& \emph{\#token} &\emph{\#rat} \\ %[0.5ex] 
 %\hline
 \hline
 RSG & 8399 & 300 & 9.92 & 0.31 \\ 
 %\hline
 AC & 11512 & 297 &9.87 & 0.15 \\
 %\hline
 BPP & 7997 & 296 & 9.95 & 0.21 \\
 %\hline
 BOA & 2706 & 309 & 9.92 & 0.2 \\
 \hline
 MR & 1800 & 32.6 & 21.2 & 8.0 \\ %[1ex] 
 %\hline
\end{tabular}
% bcw: possibly we'll remove the abbreviations, as these are provvided in the text.
\caption{Dataset characteristics. \emph{N} is the number of instances, \emph{\#sen} is the average sentence count, \emph{\#token} is the average token per-sentence count and \emph{\#rat} is the average number of rationales per document.} %For MR, we run 9-fold cross validation. Each fold has 200 documents, and each training iteration, we train on 1600 documents, and test on 200 documents.
%Ye_11: #sen is the average sentence count per document. 

%}%RSG: Random sequence generation, AC: Allocation concealment, BPP: Blinding of participants and personnel, BOA: Blinding of outcome assessment, MR: movie review dataset. 
\vspace{-1.0em}
\label{Dataset}
\end{table}

\subsection{Movie Review Dataset}
\label{section:movie-review-dataset}

We also ran experiments on a movie review (MR) dataset with accompanying rationales. \newcite{pang2004sentimental} developed and published the original version of this dataset, which comprises 1000 positive and 1000 negative movie reviews from the Internet Movie Database (IMDB).\footnote{\url{http://www.imdb.com/}} ~\newcite{zaidan2007using} then augmented this dataset by adding rationales corresponding to the binary classifications for 1800 documents, leaving the remaining 200 for testing. Because 200 documents is a modest test sample size, we ran 9-fold cross validation on the 1800 annotated documents (each fold comprising 200 documents). 
The rationales, as originally marked in this dataset, were sub-sentential snippets; for the purposes of our model, we considered the entire sentences containing the marked snippets as rationales. 

%\footnote{For MR, we run 9-fold cross validation. Each fold has 200 documents, and each training iteration, we train on 1600 documents, and test on 200 documents.}
%We'll train our model on the 1800 documents, and test it on the 200 documents. 

\section{Experimental Setup}
\label{section:experiment-setup}

%%% bcw -- pasting back in from above -- @TODO massage and again emphasize why these are the relevant baselines! this is make or break

\subsection{Baselines}
\label{section:baselines}

% punchline: primarily aim to show that rationales can improve the already very strong performance of CNN

We compare against several baselines to assess the advantages of directly incorporating rationale-level supervision into the proposed CNN architecture. 
We describe these below.

%\subsubsection{SVMs}
\vspace{.25em}

\noindent {\bf SVMs}. We evaluated a few variants of linear Support Vector Machines (SVMs). These rely on sparse representations of text. We consider variants that exploit uni- and bi-grams; we refer to these as \emph{uni-SVM} and \emph{bi-SVM}, respectively. We also re-implemented the rationale augmented SVM (\emph{RA-SVM}) proposed by ~\newcite{zaidan2007using}, described in Section \ref{section:related-work}. %In this variant, the standard SVM objective is modified to encode a preference for parameter values that result in instances that include manually annotated rationales being more confidently classified than `pseudo'-instances from which these rationales have been stripped. 

For the RoB dataset, we also compare to a recently proposed multi-task SVM (MT-SVM) model developed specifically for these RoB datasets~\cite{marshall2015automating,marshall2016robotreviewer}. This model exploits the intuition that the risks of bias across the domains codified in the aforementioned Cochrane RoB tool will likely be correlated. That is, if we know that a study exhibits a high risk of bias for one domain, then it seems reasonable to assume it is at an elevated risk for the remaining domains. Furthermore, \newcite{marshall2016robotreviewer} include rationale-level supervision by first training a (multi-task) \emph{sentence-level} model to identify sentences likely to support RoB assessments in the respective domains. Special features extracted from these predicted rationales are then activated in the \emph{document-level} model, informing the final classification. This model is the state-of-the-art on this task.

%The baseline SVMs we'll use is Uni-gram SVM (Uni-SVM) and Bi-gram SVM (Bi-SVM). Then we re-implement rationale augmented SVM (RA-SVM) in~\cite{zaidan2007using}, which is a variant of SVM built upon Uni-SVM incorporating rational information. RA-SVM is observed to consistently outperfm Uni-SVM. We use an independent development set to optimize $\mu$, $C$, and $C_{contrast}$ in their paper. For RoB dataset, we also report the state of the art result using multi-task SVM (MT-SVM)~\cite{marshall2016robotreviewer}. 

%\subsubsection{CNNs}

\vspace{.25em}

%Our main contribution is an augmentation of the CNN architecture for text classification to incorporate rationale-level supervision. Thus 
\noindent {\bf CNNs}. We compare against several baseline CNN variants to demonstrate the advantages of our approach. We emphasize that our focus in this work is \textbf{not} to explore how to induce generally `better' document vector representations -- this question has been addressed at length elsewhere, e.g., ~\cite{le2014distributed,jozefowicz2015empirical,tang2015document,yang2016hierarchical}. 

Rather, the main contribution here is an augmentation of CNNs for text classification to capitalize on rationale-level supervision, thus improving performance and enhancing interpretability. This informed our choice of baseline CNN variants: standard CNN \cite{kim2014convolutional}, Doc-CNN (described above) and AT-CNN (also described above) that capitalizes on an (unsupervised) attention mechanism at the sentence level, described in Section \ref{section:AT-CNN}.\footnote{We also experimented briefly with LSTM and GRU (Gated Recurrent Unit) models, but found that simple CNN performed better than these. Moreover, CNNs are relatively robust and less sensitive to hyper-parameter selection.}

%So we only choose CNN and Doc-CNN as the baseline for comparison. 
%\footnote{We actually found that simple CNN model perform much better than (hierachical) LSTM and GRU model on our datasets, and CNN is relatively more robust and less sensitive to hyper-parameters compared to LSTM and GRU.}

\begin{comment}
\subsection{SVM}
For SVM classifier, we use linear kernel and Stochastic Gradient Descent (SGD) as the optimization method. 

\noindent
\mathbf{RA-SVM} We follow exactly the same procedure in \cite{zaidan2007using}: we use an independent development set to optimize $\mu$, $C$, and $C_{contrast}$ in their paper. 

\subsection{CNN}
\noindent
\mathbf{CNN} We use 3 different filter heights: 3, 4 and 5, and for each filter height, we use 100 feature maps. We set the dropout rate as 0.5, and max norm constraint on the softmax layer as 3. 

\noindent
\mathbf{Doc-CNN} The setup is the same as CNN, except we use 20 feature maps for each filter height, so we end up with 60 feature maps for each sentence in the document, as well as for the document representation. 
\end{comment}

% bcw: from above
% For both CNN and Doc-CNN, we'll use back-propagation algorithm~\cite{lecun2015deep} and train the word embedding layer, the convolution layer and the softmax layer. 

\subsection{Implementation/Hyper-Parameter Details}
\label{section:configurations}

%We now describe the model configuration used for all CNNs. 

% bcw: ye -- is this right? or did we do nested CV to find?
% bcw: can we also include the number of features we used and the encoding? (e.g, `we used the top XXX most frequently occuring uni-/bi-grams and binary bag-of-words')

%\subsubsubsection{SVM-based models}

\begin{table*}
\centering
\small 
 \begin{tabular}{l c c c c} 
 %\hline
Method & RSG & AC & BPP & BOA \\[0.5ex]
\hline %\hline
\emph{Uni-SVM} & 72.16 & 72.81  & 72.80 & 65.85 \\
\emph{Bi-SVM} & 74.82 &73.62 & 75.13& 67.29 \\
\emph{RA-SVM} & 72.54 &74.11 & 75.15 & 66.29  \\
\emph{MT-SVM} & 76.15 & 74.03 & 76.33 & 67.50 \\
%Non-MT-SVM & 75.94 & 74.74 & 76.23 & 67.66 \\
\emph{CNN} & 72.50 (72.22, 72.65) & 72.16 (71.49, 72.93)& 75.03 (74.16, 75.44) & 63.76 (63.12, 64.15) \\
\emph{Doc-CNN} & 72.60 (72.43, 72.90) &72.92 (72.19, 73.48)  & 74.24 (74.03, 74.38) & 63.64 (63.23, 64.37)\\
\emph{AT-CNN} & 74.14 (73.40, 74.58) & 73.66 (73.12, 73.92)& 74.29 (74.09, 74.74) & 63.34 (63.21, 63.49)\\
\emph{RA-CNN} & \textbf{77.42 (77.33, 77.59)} & \textbf{76.14 (75.89, 76.29)} & \textbf{76.47 (76.15, 76.75)} & \textbf{69.67 (69.33, 69.93)}\\
\hline
\emph{Human} & 85.00 & 80.00 & 78.10 & 83.20  \\
\hline
\end{tabular}
\caption{Accuracies on the four RoB datasets. Uni-SVM: unigram SVM, Bi-SVM: Bigram SVM, RA-SVM: Rationale-augmented SVM \protect\cite{zaidan2007using}, MT-SVM: a multi-task SVM model specifically designed for the RoB task, which also exploits the available sentence supervision  \protect\cite{marshall2016robotreviewer}. We also report an estimate of human-level performance, as calculated using subsets of the data for each domain that were assessed by two experts (one was arbitrarily assumed to be correct). We report these numbers for reference; they are not directly comparable to the cross-fold estimates reported for the models.}
\label{RoB_perf}
\vspace{-.75em}
\end{table*}

 \begin{comment}
 \begin{figure}%[t]
%\centering
\includegraphics[width=0.5\textwidth]{RoB_results-new-3.pdf}
\caption{Accuracies on the four RoB datasets; RA-CNN uniformly outperforms the baselines.}%(For all the CNN models, we give error bar plots. )}
\vspace{-1em}
\label{figure:ROB_result}
\end{figure}
\end{comment}}

\noindent {\bf Sentence splitting.} To split the documents from all datasets into sentences for consumption by our Doc-CNN and RA-CNN models, we used the Natural Language Toolkit (NLTK)\footnote{http://www.nltk.org/api/nltk.tokenize.html} sentence splitter.  

\vspace{.25em}

\noindent {\bf SVM-based models.} We kept the 50,000 most frequently occurring features in each dataset. For estimation we used SGD. We tuned the $C$ hyper-parameter using nested development sets. For the RA-SVM, we additionally tuned the $\mu$ and $C_{contrast}$ parameters, as per \newcite{zaidan2007using}.

\vspace{.25em}

\noindent {\bf CNN-based models.}  For all models and datasets we initialized word embeddings to pre-trained vectors fit via Word2Vec. For the movie reviews dataset these were 300-dimensional and trained on Google News.\footnote{https://code.google.com/archive/p/word2vec/} For the RoB datasets, these were 200-dimensional and trained on biomedical texts in PubMed/PubMed Central \cite{moendistributional}.\footnote{http://bio.nlplab.org/}

%Word2vec with dimension 200~\cite{moendistributional} on RoB dataset. 

Training proceeded as follows. We first extracted all sentences from all documents in the training data. The distribution of sentence types is highly imbalanced (nearly all are neutral). Therefore, we downsampled sentences before each epoch, so that sentence classes were equally represented. After training on sentence-level supervision, we moved to document-level model fitting. For this we initialized embedding and convolution layer parameters to the estimates from the preceding sentence-level training step (though these were further tuned to optimize the document-level objective).
%Ye_11: I don't do downsampling for each mini-batch. 
%Ye_11: I do a randomly downsampling on the whole sentence level dataset at each epoch. (So in each epoch, the sentences might be different) And then I construct mini-batch. 

%, i.e., sets comprising an equal number of neutral, positive rationale and negative rationale sentences.

%And when we train on these sentences, for each mini-batch (size 10), the number of non-rationale sentences are much larger than that of positive and negative rationales, which would cause imbalance problem. To overcome this, we keep all the rationales, and downsample the non-rationales to make the number of positive rationales, negative rationales and non-rationales equal in one mini-batch. 

%After training on the sentence-level supervision, we moved to document-level model fitting, using mini-batch sizes of 50. We initialized embedding and convolution layer parameters to the estimates from the preceding sentence-level training step (though these were further tuned to optimize the document-level objective). 

% bcw: please provide the range we explore for the dropout below
%Ye_11: 0.0-0.9
For RA-CNN, we tuned the dropout rate (range: 0-.9) applied at the sentence vector level on each training fold (using a subset of the training data as a validation set) during the document level training phase. 
%Ye_19: I add we tune sentence level dropout rate only during the document level training step. In sentence level, it is not important, just as we found in our sensitivity analysis paper. But when we use the probability of sentence as the weight, and sum sentence vectors up to document vector, it is important. 
Anecdotally, we found this has a greater effect than the other model hyperparameters, which we thus set after a small informal process of experimentation on a subset of the data. Specifically,  we fixed the dropout rate at the document level to 0.5, and we used 3 different filter heights: 3, 4 and 5, following \cite{zhang2015}. For each filter height, we used 100 feature maps for the baseline CNN, and 20 for all the other CNN variants. %(which have a greater number of parameters).

For parameter estimation we used ADADELTA \cite{zeiler2012adadelta}, mini-batches of size 50, and an early stopping strategy (using a validation set). 
%We tuned all hyperparameters of CNN models on nested development sets (i.e., subsets of available training data). For sentence vector dropout rate, we tuned it within a small validation set on the training folds, for each test fold, so it depends on the training data. 

\section{Results and Discussion}
\label{section:results}

% bcw: what is k below? I presume 5??
\subsection{Quantitative Results}

For all CNN models, we replicated experiments 5 times, where each replication constituted 5-fold and 9-fold CV
respectively the RoB and the movies datasets, respectively. 
%Ye_31: mention CV information
We report the mean and observed ranges in accuracy across these 5 replications for these models, because attributes of the model (notably, dropout) and the estimation procedure render model fitting stochastic~\cite{zhang2015}. We do not report ranges for SVM-based models because the variance inherent in the estimation procedure is much lower for these simpler, linear models. 

Results on the RoB datasets and the movies dataset are shown in Tables \ref{RoB_perf} and Table \ref{Movie_Perf}, respectively. 
%We also display these graphically in Figures \ref{figure:ROB_result} and \ref{figure:movie_result}. 
RA-CNN consistently outperforms all of the baseline models, across all five datasets. We also observe that CNN/Doc-CNN do not necessarily improve over the results achieved by SVM-based models, which prove to be strong baselines for longer document classification. 
%Ye_09: Just to clarify more. 
This differs from previous comparisons in the context of classifying shorter texts. In particular, in previous work  ~\cite{zhang2015} we observed that CNN outperforms SVM uniformly on sentence classification tasks (the average sentence-length in these datasets was about 10). In contrast, in the datasets we consider in the present paper, documents often comprise hundreds of sentences, each in turn containing multiple words. We believe that it is in these cases that explicitly modeling which sentences are most important will result in the greatest performance gains, and this aligns with our empirical results.%The strong performance of RA-CNN suggests that taking into account document structure is key when applying CNN to longer texts.
%Ye_13: I guess this might answer reviewer 2's question? 
% bcw_14: dropping for now because I think too vague. Added a different statement above instead for now
%Ye_14: I think Doc-CNN is not strong compared to SVM, and it's not universally better than pure CNN. 
%Ye_14: The reviewer asks to analyze the relation between the performance and the length of document, but I don't think this point is important, maybe just drop what I write. 
%For RA-CNN, we don't observe any correlation between the performance and the document length or sentence length. 

Another observation is that AT-CNN does often improve performance over vanilla variants of CNN (i.e., without attention), especially on the RoB datasets, probably because these comprise longer documents. However, as one might expect, RA-CNN clearly outperforms AT-CNN by exploiting rationale-level supervision directly.
%Ye_11: AT-CNN does not always improve performance, and sometimes ever hurt. 
% bcw_11: I softened, but on RoB datasets it does seem to help overall (exception: BOA).. anyway your point is taken
And by exploiting rationale information directly, RA-CNN is able to consistently perform better than baseline CNN and SVM model variants. Indeed, we find that RA-CNN outperformed MT-SVM on all of the RoB datasets, and this was accomplished without exploiting cross-domain correlations (i.e., without multi-task learning).

\begin{table}
\small
\centering
 \begin{tabular}{l c} 
 Method & Accuracy \\
 \hline
\emph{Uni-SVM} & 86.44\\
\emph{Bi-SVM} & 86.94\\
\emph{RA-SVM} & 88.89\\
\emph{CNN} & 85.59 (85.27, 86.17) \\
\emph{Doc-CNN} & 87.14 (86.70, 87.60)\\
\emph{AT-CNN} & 86.69 (86.28, 87.17) \\
\emph{RA-CNN} & \textbf{90.43 (90.11, 91.00)}\\
\hline
\end{tabular}
\caption{Accuracies on the movie review dataset.}

\label{Movie_Perf}
\vspace{-1.25em}
\end{table}

\begin{comment}
\begin{figure}%[t]
\centering
\includegraphics[width=0.5\textwidth]{movie_results-new-3.pdf}
\vspace{-1em}
\caption{Accuracies achieved on the movies dataset. RA-CNN outperforms all baseline models.}%Accuracy of different models on movie review dataset (For all the CNN models, we give error bar plots. )}
\label{figure:movie_result}
\vspace{-1em}
\end{figure}
\end{comment}

\vspace{-.5em}
\subsection{Qualitative Results: Illustrative Rationales}
\vspace{-.1em}
\label{section:predicted-rationales}

In addition to realizing superior classification performance, RA-CNN also provides \emph{explainable} categorizations. The model can provide the highest scoring rationales (ranked by max$\{p_\text{pos},p_\text{neg}\}$) for any given target instance, which in turn -- by construction -- are those that most influenced the final document classification. %To illustrate this, we display rationales automatically extracted from a sample of correctly classified test documents from the RoB datasets in Table \ref{table:rationales}. 

For example, a sample positive rationale supporting a correct designation of a study as being at low risk of bias with respect to blinding of outcomes assessment reads simply \emph{The study was performed double blind}. An example rationale extracted for a study (correctly) deemed at high risk of bias, meanwhile, reads \emph{as the present study is retrospective, there is a risk that the woman did not properly recall how and what they experienced ...}.

Turning to the movie reviews dataset, an example rationale extracted from a glowing review of `Goodfellas' (correctly classified as positive) reads \emph{this cinematic gem deserves its rightful place among the best films of 1990s}. While a rationale extracted from an unfavorable review of `The English Patient' asserts that \emph{the only redeeming qualities about this film are the fine acting of Fiennes and Dafoe and the beautiful desert cinematography}. 

In each of these cases, the extracted rationales directly support the respective classifications. This provides direct, meaningful insight into the automated classifications, an important benefit for neural models, which are often seen as opaque.

\section{Conclusions}
%\vspace{-.5em}
\label{section:conclusions}

We developed a new model (RA-CNN) for text classification that extends the CNN architecture to directly exploit \emph{rationales} when available. We showed that this model outperforms several strong, relevant baselines across five datasets, including vanilla and hierarchical CNN variants, and a CNN model equipped with an attention mechanism. Moreover, RA-CNN automatically provides explanations for classifications made at test time, thus providing interpretability. 
%Ye_09

Moving forward, we plan to explore additional mechanisms for exploiting supervision at lower levels in neural architectures. Furthermore, we believe an alternative approach may be a hybrid of the AT-CNN and RA-CNN models, wherein an auxiliary loss might be incurred when the attention mechanism output disagrees with the available direct supervision on sentences.

%The future work might be infusing the loss of the AT-CNN architecture with a measure of agreement between the attention distribution and rationale annotations. 

% BCW: maybe close with future directions...

%This is achieved via a hierarchical approach in which documents are represented by a weighted sum of sentence-level embeddings, where weights reflect the estimated probabilities of sentences providing supporting assessments. 

%hat can utilize the rationales. It combines sentence level CNN and document level CNN , and consistently outperforms the baseline CNN models. It can highlight the sentences that support the label of the test document, which is not available in traditional rationale augmented SVM. It is flexible enough that CNN could be replaced with other neural models.

%%
% TODO comment out for submission!
%Ye_09: Update the acknowledgements? BCW: I updated!
\section*{Acknowledgments}

Research reported in this article was supported by the National Library of Medicine (NLM) of the National Institutes of Health (NIH) under award number R01LM012086. The content is solely the responsibility of the authors and does not necessarily represent the official views of the National Institutes of Health. This work was also made possible by the support of the Texas Advanced Computer Center (TACC) at UT Austin.

%Do not number the acknowledgment section.
%\newpage
\bibliography{emnlp2016}
\bibliographystyle{emnlp2016}

\end{document}